\documentclass[10pt,twocolumn,letterpaper]{article}

\usepackage[pagenumbers]{cvpr}

\usepackage{graphicx}
\usepackage{amsmath}
\usepackage{amssymb}
\usepackage{booktabs}
\usepackage{arydshln}

\usepackage[pagebackref,breaklinks,colorlinks]{hyperref}

\usepackage[capitalize]{cleveref}
\crefname{section}{Sec.}{Secs.}
\Crefname{section}{Section}{Sections}
\Crefname{table}{Table}{Tables}
\crefname{table}{Tab.}{Tabs.}

\begin{document}

\title{Language-Based Depth Hints for Monocular Depth Estimation}

\author{Dylan Auty, Krystian Mikolajczyk\\
Imperial College London\\
London, SW7 2AZ, United Kingdom\\
{\tt\small \{dylan.auty12,k.mikolajczyk\}@imperial.ac.uk}
}
\maketitle

\begin{abstract}

Monocular depth estimation (MDE) is inherently ambiguous, as a given image may result from many different 3D scenes and vice versa. To resolve this ambiguity, an MDE system must make assumptions about the most likely 3D scenes for a given input. These assumptions can be either explicit or implicit.
In this work, we demonstrate the use of natural language as a source of an explicit prior about the structure of the world. The assumption is made that human language encodes the likely distribution in depth-space of various objects. We first show that a language model encodes this implicit bias during training, and that it can be extracted using a very simple learned approach. We then show that this prediction can be provided as an explicit source of assumption to an MDE system, using an off-the-shelf instance segmentation model that provides the labels used as the input to the language model.
We demonstrate the performance of our method on the NYUD2 dataset, showing improvement compared to the baseline and to random controls\footnote{This work was originally done in June 2022.}.

\end{abstract}
\section{Introduction}
\label{sec:introduction}

Monocular depth estimation (MDE) from a single image is an ill-posed problem due to its inherent ambiguity: a given 2D image may have been taken from infinitely many 3D scenes, and this ambiguity may not be resolved as information is lost in the process of taking this image.
Nonetheless, both biological vision systems and deep learning MDE methods have achieved success in this task, as contextual cues and assumptions about the world exist that can be used to reduce the solution space and make this problem tractable. Depending on the system used, the means of ingesting and interpreting these contextual cues or assumptions differ: some model architectures are designed explicitly around certain assumptions, while other methods expect that this information is absorbed implicitly during the training process.

The use of cues and assumptions in biological MDE is well established \cite{hershenson_pictorial_1998, wagner_barn_1991, nagata_depth_2012, harkness_chameleons_1977, sousa_judging_2011}. In this work, we build on this idea further.
Human language exists to convey human ideas, which in turn must contain human biases; therefore, we hypothesise that a sufficiently large corpus of human language will implicitly contain useful information about the human biases that are used for human monocular depth perception. Recent work \cite{henlein_what_2022} has demonstrated that BERT \cite{devlin_bert_2019} encodes relationships between objects and the rooms they most frequently occur in, or verbs and the objects with which they are most often associated. In this work, we extend this analysis directly to assumptions on the most probable distribution of depth for a given object, and demonstrate a method to leverage this information to improve the performance of existing MDE systems on the NYUD2 dataset.

Our main contributions are:
\begin{enumerate}
    \item We demonstrate the existence of an inherent depth bias in BERT, that relates object names to their likely depth values, and present a method of extracting this information,
    \item We show a novel method of incorporating this bias to an existing MDE system in an adaptable way,
    \item We achieve improved performance compared to the baseline and control methods on the NYUD2 dataset.
\end{enumerate}
\section{Background}
\label{sec:background}

\noindent\textbf{Monocular Depth Estimation.} Monocular depth estimation methods generally treat the problem as a dense image to image mapping. They often consist of a convolutional encoder and decoder, and frequently make use of encoders that have been pretrained on ImageNet \cite{deng_imagenet:_2009} such as ResNet \cite{he_deep_2016}, VGG \cite{simonyan_very_2015}, or EfficientNet \cite{tan_efficientnet_2019}. When dense features have been extracted from these encoders, they are then upsampled or decoded to provide an output depth map.

Previous SOTA methods have expanded on this basic template in different ways. \cite{eigen_depth_2014, eigen_predicting_2015, lee_big_2019} use a multi-scale prediction approach. \cite{lee_big_2019} make the explicit assumption that the world is (locally) planar, and also incorporate a multi-scale coarse-to-fine idea. Some methods co-predict depth with other tasks such as semantic segmentation or surface normals \cite{ramamonjisoa_sharpnet:_2019, jiao_look_2018, bai_monocular_2019}, sharing an encoder or sharing weights between the different decoder heads, or enforcing consistency between the predictions in the loss function.

Other methods have rephrased the problem: rather than being a regression problem with a scalar output, \cite{fu_deep_2018} treated it as a classification problem, where the model must assign each pixel of the input image to a depth "bin". This method performed well but produced artifacts due to the discrete nature of the output. AdaBins \cite{bhat_adabins_2020} expanded on this work significantly: rather than using uniform bins, a transformer \cite{vaswani_attention_2017} based model was used to adaptively change the distribution and width of the bins according to the input image, and instead of directly classifying pixels to bins, the probability vectors were used to smoothly interpolate between the adaptive bin centres. \cite{li_binsformer_nodate} improve on the adaptive binning concept by incorporating intermediate features as part of the input to the adaptive binning transformer and by incorporating an auxiliary scene classification task to encourage the model to learn global semantic information about the scene. \cite{li_depthformer_2022} use both a transformer and a traditional convolutional encoder to extract image features, before merging them and decoding to a depth map using a transformer decoder.

\noindent\textbf{Language and BERT.} 
BERT \cite{devlin_bert_2019} is a transformer-based \cite{vaswani_attention_2017} language model introduced in 2019. Whereas older language modelling/word embedding methods such as Word2Vec \cite{mikolov_efficient_2013} or GloVe \cite{pennington_glove_2014} produce embeddings that are static, BERT is highly sensitive to sequential patterns, and produces \textit{contextual} word embeddings that change based on the sentence in which the input words are used. Because of this property, it is better than static models at many common language tasks, for instance the disambiguation of polysemous words (e.g. the word ``bank" in the sentence ``The bank robber jumped off the river bank").

Following the invention of BERT, further work has been done to investigate the nature of the information that it is able to learn from its training corpora. BERT embeddings have been shown to encode whole syntax trees \cite{hewitt_structural_2019} and knowledge graphs \cite{wang_language_2020}, and to contain a degree of semantic information \cite{ettinger_what_2020}. It also contains some knowledge about the world in general: using fill-in-the-gaps prompts, and without fine-tuning, BERT already contains some relational knowledge between certain entities \cite{petroni_language_2019}.

\noindent\textbf{Language and depth.} Language and depth is a relatively new area of research. Intuitively, a language model exists to encode human language; if human language in turn encodes human ideas, then it would follow that language models encode human biases. \cite{henlein_what_2022} showed that BERT embeddings contain information about the likely distribution of objects across different rooms, the relationship between parts of objects and the objects they are likely part of, and the likely target objects for a given verb. \cite{auty_monocular_2022} show that certain biological depth cues are useful for depth estimation, and make use of GloVe embeddings as a means of introducing weak world knowledge to the model. CLIP \cite{radford_learning_2021} is a transformer-based model that, given enough data, is able to correlate high-level visual concepts with language; this has been used for image generation from language \cite{ramesh_hierarchical_2022}, and has more recently been applied to zero-shot depth estimation \cite{zhang_can_2022} by correlating image features with the features of depth-related sentences. The downside of this method is the enormous quantity of data and computation time required for training CLIP-like models. BERT is less computationally intensive to train and use, achieves excellent performance on language tasks, and a significant body of research has already been undertaken in the literature investigating its properties. For these reasons, our work focuses on the use of BERT.

\section{Method}
\label{sec:method}

\begin{figure*}
    \centering
    \includegraphics[width=\textwidth]{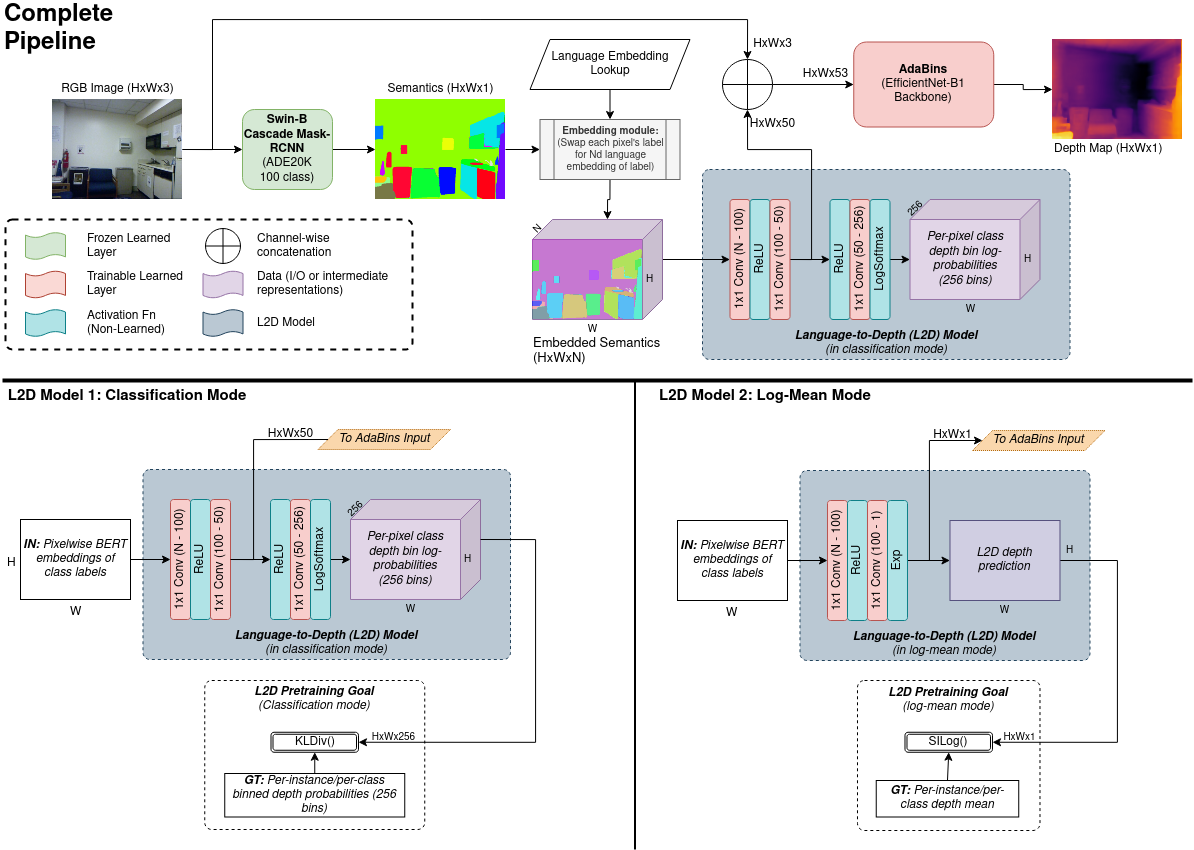}
    \caption{\textbf{Top}: The complete MDE pipeline of our method, showing the L2D model in the classification configuration. The L2D model used is pretrained separately, then the pretrained checkpoint is loaded at the start of training of the full MDE pipeline. \textbf{Bottom:} The two configurations experimented with for the L2D model: "classification" mode and "log-mean" mode (see sec. \ref{sec:l2d-model}).}
    \label{fig:pipeline-1}
\end{figure*}

Building on the language-and-depth work in the literature already discussed in section \ref{sec:background}, we introduce a novel method to first extract inherent language-to-depth bias from BERT embeddings, and then to insert this bias into an MDE pipeline to enhance performance.

Following from the work of \cite{henlein_what_2022}, we extend the intuition that language contains human biases to depth: specifically, we assume that humans will use a object noun in contexts that semantically make sense for that object. This in turn will encode either explicit or implicit information about the likely depth values of an object.

While \textit{explicit} depth information (e.g. \textit{"\{Object\} is nearby/in the distance"}) may or may not be common in language, it is hypothesised that \textit{implicit} depth information may be more common. This can include relative depths or even other properties of an object that relate to depth, such as size. Aggregated across a large corpus of natural language, slightly-biased utterances such as these would sum to an overall bias for a variety of objects. It is this aggregate bias that our method extracts and exploits for MDE.

Our model pipeline is shown in figure \ref{fig:pipeline-1}. It consists of two main components: the RGB-to-depth model, and the language-to-depth (L2D) model. To bridge the gap between the input image and the language-to-depth model, we make use of an off-the-shelf instance segmentation network to provide per-pixel labels. The model used for this purpose is a Cascade Mask-RCNN with a Swin-B backbone \cite{liu_swin_2021} trained as an instance segmentation model on the ADE20K dataset's 100-class "Places Challenge" subset \cite{zhou_scene_2017, zhou_semantic_2018}. The instance segmentation model is frozen during training. The L2D models that are used are first described in section \ref{sec:l2d-model}, and then the RGB-to-depth model is described in section \ref{sec:mde-model}.

\subsection{Language-To-Depth (L2D) Model}
\label{sec:l2d-model}
The language to depth (L2D) model architectures build on the findings of \cite{henlein_what_2022}, who showed that while a spatial bias of the relationship between objects and the rooms they are likely to occur in is encoded in BERT embeddings, this relationship is too complex to be extracted using simple cosine similarity or distance measures, and must instead be extracted using a simple feed-forward network with a single hidden layer of 100 neurons. Similarly, the L2D model used in our method to extract the inherent depth bias in BERT is a relatively simple FFN with only 1-2 hidden layers (depending on configuration).

The two different architectures experimented with for the L2D model are shown at the bottom of figure \ref{fig:pipeline-1}. The first architecture (the "classification" mode of the L2D model) was pretrained with a depth-bin classification objective: the range of possible depth values (0-10m) is divided into 256 bins, and the model must map the language embedding to probabilities across those 256 bins. The second architecture (the "log-mean" mode of the L2D model) was pretrained with a direct depth-mean prediction objective: the model must map a semantic class's language embedding to the mean depth of that class, as it appears in the pretraining dataset.

When used as part of the complete pipeline shown at the top of figure \ref{fig:pipeline-1}, the inputs to the L2D model are the pixelwise semantic labels predicted by the Cascade Mask-RCNN, embedded by the language model being used. In the case of the "log-mean" L2D model configuration, the pixelwise depth mean prediction is concatenated to the image before input to the AdaBins-B1 model. In the case of the "classification" model, the model includes more layers than in the "log-mean" configuration that compress the dimensionality of the embedding progressively, before the final 256-channel output layer; the pixelwise 50-dimensional intermediate embeddings from from the second-to-last layer are what are provided to the AdaBins-B1 model at the input.

\subsubsection{BERT Word Embedding Strategies}
\label{sec:bert-word-embedding-strategies}
Two pretrained BERT models were tested: BERT-base \cite{devlin_bert_2019}, which produces 768-dimensional embeddings, and BERT-tiny \cite{turc_well-read_2019}, a smaller version of BERT that produces 128-dimensional embeddings. BERT embeddings for a given target word are produced in one of two ways for use as inputs to the language-to-depth model.

The first approach, the "non-contextual" version, inserts the target word between the [CLS] and [SEP] tokens that BERT expects and passes it through BERT. The hidden states from the last four layers are then taken and averaged together for each token, then all average token embeddings are themselves averaged together. This method derives from experiments in the original BERT paper \cite{devlin_bert_2019} on the CoNLL-2003 Named Entity Recognition Task \cite{tjong_kim_sang_introduction_2003}, where a similar approach provides an optimal balance between embedding dimensionality and performance.

The second approach, the "contextual" version, derives from the approach of \cite{henlein_what_2022}, who in turn use the approach of \cite{may_measuring_2019}: the target words are inserted into semantically barren phrases such as \textit{"[CLS] This is a/an \{word\} [SEP]"} or \textit{"[CLS] \{Word\} is here [SEP]"}. These serve to wrap the word in the complete sentence format that will be familiar to the language model, without introducing additional semantic prompting that may distort the embedding generated. Each sentence is run through BERT, then the final hidden state for the target word's token in each of the template sentences is averaged together to give the final word embedding.

\subsubsection{L2D Pretraining Datasets}
\label{sec:pretraining-datasets}
The two pretraining data used were prepared from NYUD2, with instance and semantic annotations provided by the Swin-B Cascade Mask-RCNN used as part of the complete pipeline (see section \ref{sec:method} and fig. \ref{fig:pipeline-1}).

The first pretraining dataset is denoted "instance" or "inst" in this work: each instance of each object detected in NYUD2 is recorded as a data point. For the "classification" configuration of the L2D model, each data point is a pair of the instance's semantic label with the distribution of depth values shown in that instance presented across 256 bins across the 0-10m depth range. For the "log-mean" configuration of the L2D model, each data point is a pair of the instance's semantic label with the mean depth value in that instance. The "inst" dataset comprised a training set of 777636 and a test set of 21521 label/average depth pairs, drawn from the training and testing sets respectively of NYUD2.

The second pretraining dataset is denoted "leave-one-out" or "loo" in this work. This is a per-class dataset instead of a per-instance dataset. Each data point is a pair of semantic label, with either the average depth of all instances of that label (for "log-mean" L2D pretraining) or depth probability distribution across the range of depths (for "classification" L2D pretraining, again split across 256 bins). The "loo" contained 101 total data points, one for each class in the ADE20K-Places instance segmentation dataset and an additional one for background or unknown objects (for which the word "background" was used as a label).

In either case, only data drawn from the NYUD2 training split is used for pretraining the L2D model, to avoid contamination of the final MDE results. Pretraining on the "inst" dataset was done for 100 epochs with a batch size of 1000. Pretraining on the "leave-one-out" dataset was done for 100 epochs with a batch size of 100 (all classes except for the one left out by the leave-one-out process).

\subsection{RGB-To-Depth Model}
\label{sec:mde-model}
To provide the primary depth predictions, the architecture of AdaBins \cite{bhat_adabins_2020} was used, but modified to use an EfficientNet-B1 encoder rather than an Efficientnet-B5 \cite{tan_efficientnet_2019}. This was necessary due to a lack of available GPU RAM; the EfficientNet-B5 contains 30M parameters, compared to the 7.8M parameters of EfficientNet-B1. Naturally, this reduction in size brings a likely decrease in performance that is unfortunately necessary due to hardware limitations. This modified architecture is referred to in this paper as "AdaBins-B1", and is used as the baseline method.

The AdaBins architecture is composed of a dense encoder/decoder that produces dense features for the input image, and an adaptive binning layer. The latter is a smaller version of the vision transformer \cite{dosovitskiy_image_2021}, and is responsible for both producing adaptively sized depth-bins for the input image, and assigning each pixel probabilities of being in the resultant bins. Finally, scalar depth values are produced by interpolating between the adaptive bin centres using the assigned bin probabilities for a given pixel. The L2D models described in section \ref{sec:l2d-model} are incorporated into the RGB-to-depth model at the input stage, by concatenating pixelwise predictions or features from the L2D model with the input image (see section \ref{sec:l2d-model}).

\begin{table*}
    \centering
    \begin{tabular}{ll|llll|lll}
    Embeddings & Pre. & \multicolumn{4}{l|}{Distance metrics ($\downarrow$)} & \multicolumn{3}{l}{$\delta$ accuracy ($\uparrow$)} \\
                                         & & Abs Rel   & Sq Rel   & RMS   & RMSL   & $1.25$  & $1.25^2$ & $1.25^3$ \\ \hline
    \hline
    Random (128d)                & inst & \textbf{0.396} & \textbf{0.648} & 1.468 & \textbf{0.448} & 0.391 & \textbf{0.682} & \textbf{0.858} \\
    BERT-tiny (128d) \cite{turc_well-read_2019}  & inst & 0.401 & 0.657 & \textbf{1.465} & \textbf{0.448} & \textbf{0.392} & \textbf{0.682} & \textbf{0.858} \\
    BERT-tiny cntxt. (128d)   & inst & 0.402 & 0.659 & \textbf{1.465} & \textbf{0.448} & \textbf{0.392} & \textbf{0.682} & 0.857 \\
    \hdashline
    Random (128d)           & loo & 3.584 & 1.095 & 0.161 & 1.201 & 0.101 & 0.292 & 0.371 \\
    BERT-tiny (128d)        & loo & 3.366 & 1.004 & \textbf{0.139} & 1.080 & \textbf{0.180} & \textbf{0.326} & \textbf{0.438} \\
    BERT-tiny cntxt. (128d) & loo & \textbf{3.029} & \textbf{0.645} & 0.142 & \textbf{1.056} & 0.169 & \textbf{0.326} & 0.427 \\
    
    \hline

    Random (768d)                 & inst & 0.410 & 0.680 & \textbf{1.465} & 0.450 & \textbf{0.394} & 0.680 & 0.855 \\
    BERT-base (768d) \cite{devlin_bert_2019}             & inst & 0.404 & 0.666 & 1.466 & 0.449 & 0.393 & 0.681 & 0.856 \\
    BERT-base cntxt. (768d)    & inst & \textbf{0.398} & \textbf{0.653} & 1.467 & \textbf{0.448} & 0.392 & \textbf{0.682} & \textbf{0.858} \\
    \hdashline
    Random (768d)                 & loo & 2.743 & 0.323 & 0.120 & 1.041 & 0.135 & 0.281 & 0.404 \\
    BERT-base (768d)            & loo & 2.275 & 0.341 & 0.133 & 1.082 & 0.079 & 0.303 & 0.404 \\
    BERT-base cntxt. (768d)    & loo & \textbf{2.144} & \textbf{0.273} & \textbf{0.106} & \textbf{0.848} & \textbf{0.225} & \textbf{0.393} & \textbf{0.562} \\
    
    \end{tabular}
    \caption{
    Results of pretraining the L2D in "log-mean" configuration (see section \ref{sec:l2d-model}) for 100 epochs, on either the "inst" or the "leave-one-out" ("loo") preparations of NYUD2. The best results per embedding dimensionality and pretraining dataset are shown in bold. "Cntxt." denotes contextual BERT embeddings.
    }
    \label{tab:l2d-log-mean}
\end{table*}

\subsection{Loss function}
\label{sec:loss-function}
The loss function used is the scale-invariant log loss of AdaBins \cite{bhat_adabins_2020}. If the ground-truth and predicted depth values for pixel $i$ are denoted by $d^*_i$ and $d_i$ respectively, the scale invariant log (SILog) loss is given by\footnote{Note that this formulation matches the version implemented in the official code of \cite{bhat_adabins_2020}, but differs slightly from that presented in the paper.}:

\begin{equation}
\label{eq:loss_silog}
    \mathcal{L}_{SILog} = 10 \sqrt{\frac{1}{N}\displaystyle\sum_{i \in N} g_{i}^2 + \frac{0.15}{N^2}(\displaystyle\sum_{i \in N} g_{i})^2}
\end{equation}

where $g_{s_i} = log(d_{i}) - log(d^*_{i})$ and $N$ is the total number of pixels with valid depth values.

When pretraining the L2D model in "log-mean" mode (as described in section \ref{sec:l2d-model}) is performed, the SILog Loss is used between the predicted mean depths and the ground-truth mean depths. Instead of operating over pixels, as previously described, the loss function operates over each data point in the pretraining batch.

When pretraining the L2D model in "classification" mode, the Kullback-Liebler divergence (KLDiv) is used to find the degree to which the predicted and ground-truth depth distributions differ. It is defined as:
\begin{equation}
\label{eq:loss_kldiv}
    \mathcal{L}_{KLDiv} = \sum_{i \in N}D(i)log \left( \frac{D(i)}{D^*(i)} \right)
\end{equation}

Where $D$ and $D^*$ represent the predicted and ground-truth depth distributions respectively, represented as probabilities across 256-bins, and $i$ indicates a particular data point out of the entire batch $N$. Due to an implementation detail of the PyTorch library, in practice log-probabilities are predicted instead of probabilities.

\section{Experiments}
\label{sec:experiments}

\begin{table*}
    \centering
    \small
    \begin{tabular}{lll|llll|lll}
    Embeddings & L2D & Pre. & \multicolumn{4}{l|}{Distance metrics ($\downarrow$)} & \multicolumn{3}{l}{$\delta$ accuracy ($\uparrow$)} \\
                                         & & & Abs Rel   & Sq Rel   & RMS   & RMSL   & $1.25$  & $1.25^2$ & $1.25^3$ \\ \hline
    \hline
    \multicolumn{3}{c|}{Baseline (AdaBins-B1 only)} & 0.123 & 0.076 & 0.413 & 0.151 & 0.864 & 0.976 & 0.995 \\
    \hline
    
    Random (128d)            & log-mean & inst & 0.116 & 0.068 & 0.393 & 0.143 & 0.879 & 0.981 & \textcolor{red}{\textbf{0.996}} \\
    BERT-tiny (128d) \cite{turc_well-read_2019} & log-mean & inst & 0.118 & 0.068 & 0.396 & 0.145 & 0.876 & 0.980 & \textcolor{red}{\textbf{0.996}} \\
    BERT-tiny cntxt. (128d)  & log-mean & inst & \textcolor{red}{\textbf{0.115}} & \textcolor{red}{\textbf{0.065}} & \textcolor{red}{\textbf{0.392}} & \textcolor{red}{\textbf{0.142}} & \textcolor{blue}{\textbf{0.880}} & \textcolor{red}{\textbf{0.983}} & \textcolor{red}{\textbf{0.996}} \\
    
    \hdashline
    Random (128d)                 & log-mean & loo & 0.120 & 0.072 & 0.404 & 0.147 & 0.869 & 0.979 & \textcolor{red}{\textbf{0.996}} \\
    BERT-tiny (128d)             & log-mean  & loo & \textbf{0.118} & \textbf{0.070} & \textbf{0.402} & \textbf{0.146} & \textbf{0.874} & \textbf{0.981} & \textcolor{red}{\textbf{0.996}} \\
    BERT-tiny cntxt. (128d) & log-mean & loo & 0.119 & 0.071 & 0.404 & \textbf{0.146} & 0.873 & 0.980 & 0.995 \\
    
    \hdashline
    Random (128d)                 & class. & inst & 0.117 & 0.068 & \textbf{0.394} & 0.144 & 0.876 & 0.980 & \textcolor{red}{\textbf{0.996}} \\
    Bert-tiny (128d)              & class. & inst & \textbf{0.116} & \textcolor{red}{\textbf{0.065}} & \textbf{0.394} & \textbf{0.143} & \textcolor{blue}{\textbf{0.880}} & \textbf{0.981} & \textcolor{red}{\textbf{0.996}} \\
    Bert-tiny cntxt. (128d)   & class. & inst & 0.119 & 0.070 & 0.396 & 0.145 & 0.877 & \textbf{0.981} & \textcolor{red}{\textbf{0.996}} \\
    
    \hdashline
    Random (128d)                 & class. & loo & \textbf{0.118} & \textbf{0.068} & \textbf{0.396} & \textbf{0.145} & \textbf{0.876} & \textbf{0.981} & \textcolor{red}{\textbf{0.996}} \\
    Bert-tiny (128d)              & class. & loo & 0.120 & 0.070 & 0.398 & 0.146 & 0.875 & 0.979 & 0.995 \\
    Bert-tiny cntxt. (128d)       & class. & loo & 0.119 & 0.070 & 0.398 & 0.146 & 0.873 & 0.979 & 0.995 \\
    
    \hline \hline
    
    Random (768d)                 & log-mean & inst & 0.119 & 0.069 & 0.399 & 0.145 & \textbf{0.878} & \textbf{0.981} & \textcolor{red}{\textbf{0.996}} \\
    BERT-base (768d) \cite{devlin_bert_2019} & log-mean & inst & 0.118 & 0.069 & \textbf{0.398} & \textbf{0.144} & 0.877 & \textbf{0.981} & \textbf{\textcolor{red}{\textbf{0.996}}} \\
    BERT-base cntxt. (768d)   & log-mean & inst & \textbf{0.117} & \textbf{0.068} & 0.400 & \textbf{0.144} & \textbf{0.878} & \textbf{0.981} & \textcolor{red}{\textbf{0.996}} \\
    
    \hdashline
    Random (768d)                & log-mean  & loo & 0.120 & \textbf{0.071} & 0.404 & 0.147 & 0.873 & \textbf{0.980} & \textcolor{red}{\textbf{0.996}} \\
    BERT-base (768d)              & log-mean & loo & 0.120 & \textbf{0.071} & \textbf{0.403} & \textbf{0.146} & 0.874 & 0.979 & \textcolor{red}{\textbf{0.996}} \\
    BERT-base cntxt. (768d)  & log-mean & loo & \textbf{0.118} & \textbf{0.071} & \textbf{0.403} & 0.147 & \textbf{0.875} & 0.979 & 0.995 \\
    
    \hdashline
    Random (768d)                   & class. & inst & 0.117 & 0.067 & \textcolor{red}{\textbf{0.392}} & \textcolor{blue}{\textbf{0.143}} & 0.879 & \textcolor{blue}{\textbf{0.982}} & \textcolor{red}{\textbf{0.996}} \\
    BERT-base (768d) & class. & inst & \textcolor{blue}{\textbf{0.116}} & \textcolor{blue}{\textbf{0.066}} & 0.395 & \textcolor{blue}{\textbf{0.143}} & \textcolor{red}{\textbf{0.882}} & \textcolor{blue}{\textbf{0.982}} & \textcolor{red}{\textbf{0.996}} \\
    BERT-base cntxt. (768d)   & class. & inst & \textcolor{blue}{\textbf{0.116}} & \textcolor{blue}{\textbf{0.066}} & 0.394 & \textcolor{blue}{\textbf{0.143}} & 0.879 & \textcolor{blue}{\textbf{0.982}} & \textcolor{red}{\textbf{0.996}} \\
    
    \hdashline
    Random (768d)                 & class. & loo & 0.120 & 0.070 & 0.399 & 0.146 & 0.873 & 0.979 & \textcolor{red}{\textbf{0.996}} \\
    BERT-base (768d) & class. & loo & 0.119 & 0.070 & 0.399 & 0.146 & 0.875 & 0.979 & 0.995 \\
    BERT-base cntxt. (768d)   & class. & loo & \textbf{0.118} & \textbf{0.069} & \textbf{0.398} & \textbf{0.145} & \textbf{0.876} & \textbf{0.980} & 0.995 \\
    
    \end{tabular}
    \caption{Results of our complete method (fig. \ref{fig:pipeline-1}) on NYUD2 after 25 epochs of training. Different language embedding sizes, L2D model configurations, and pretraining datasets are compared. "Cntxt." denotes contextual language embeddings. The "Pre." column denotes the L2D pretraining dataset used. Best results across all experiments in \textcolor{red}{\textbf{red}}. Best results for each embedding size in \textcolor{blue}{\textbf{blue}}. Best results per embedding size, L2D model, and L2D pretraining type in \textcolor{black}{\textbf{bold}}. More general colours take precedence (i.e. red $>$ blue $>$ bold).
    }
    \label{tab:results-1}
\end{table*}

This section details the experiments undertaken in this work. First, the depth bias inherently encoded in BERT is demonstrated using two different training methodologies. Secondly, the incorporation of this inherent bias into an MDE method is tested, and a performance improvement is found.

The metrics used for evaluation are those defined by \cite{eigen_depth_2014}: Abs relative difference (Abs or Abs Rel): $\frac{1}{T}\sum_{i=1}^{T} \frac{|d_i - d_i^*|}{d_i^*}$, Squared relative difference (Sq or Sq Rel): $\frac{1}{T}\sum_{i=1}^{T} \frac{\|d_i - d_i^*\|}{d_i^*}$, lin. RMSE (RMS): $\sqrt{\frac{1}{T}\sum_{i=0}^{T}\|d_i - d_i^*\|^2}$, log RMSE (RMSL): $\sqrt{\frac{1}{T}\sum_{i=0}^{T}\|log(d_i) - log(d_i^*)\|^2}$, and the threshold accuracy $\delta_n$: \(\%\) of \(d_i\) s.t. $max(\frac{d_i}{d_i^*}, \frac{d_i^*}{d_i}) = \delta < thr$, where $\delta_n$ denotes that $thr = 1.25^n$ (we use $n \in \{1, 2, 3\}$).

\subsection{Confirmation of BERT Depth Bias}
\label{sec:l2d-pretraining}

To verify the existence of depth-bias in BERT, the L2D model was trained in isolation in the "log-mean" configuration described in section \ref{sec:l2d-model}, using either the "inst" or the "leave-one-out" ("loo") preparations of NYUD2 described in section \ref{sec:pretraining-datasets}.

When using the "leave-one-out" configuration, a new model was trained for each target class. This model was trained for 100 epochs on all other classes, and then the prediction was made at the end of training on the left-out class. When models had been trained for each of the left-out classes in the dataset and predictions taken for each, error metrics were evaluated on these saved predictions and aggregated to produce the results used in this section. When using a "leave-one-out"-pretrained model as part of the main model shown in figure \ref{fig:pipeline-1}, the saved predictions for a given class were used (equivalent to using one frozen L2D model per class).

The results of this pretraining are shown in table \ref{tab:l2d-log-mean}, comparing the contextual and non-contextual BERT embedding strategies as described in section \ref{sec:bert-word-embedding-strategies} for each pretraining dataset. Also shown are results from control models trained with random "embeddings" of equivalent dimensionality to the BERT embeddings used, where the "embedding" is constant throughout training for a given class but is composed of numbers uniformly distributed in the half-open interval $[0, 1)$. In this way, the performance due to the \textit{distinctiveness} between classes may be distinguished from that due to the \textit{semantic world-knowledge and depth bias} extracted from the BERT embeddings themselves.

It can be seen that the "leave-one-out" pretraining dataset produced more distinctive improvements compared to random embeddings, particularly in the case of the smaller BERT-tiny embeddings. Further to this, the results using the higher-dimensional BERT-base embeddings were generally better than an equivalent experiment with smaller embeddings.

\subsection{Use of BERT Bias for MDE (Complete Pipeline)}
\label{sec:use-of-bert-bias-for-mde}

Having pretrained L2D models to extract the depth-bias information inherently embedded in BERT's language embeddings, these models were then incorporated into the complete pipeline with AdaBins-B1 as described in sections \ref{sec:l2d-model} and \ref{sec:mde-model}.
Pretraining was on either the "inst" or "leave-one-out" ("loo") datasets, as described in section \ref{sec:pretraining-datasets}. The "inst"-pretrained L2D model was permitted to learn during training of the complete pipeline. The "leave-one-out"-pretrained ensemble of L2D models were frozen (in practice, the predictions for each class were recorded at the end of training and used in a look up table).

The results are shown in table \ref{tab:results-1}. It may be seen that, regardless of the embedding used or the manner of training used for the L2D model, an improvement in performance is seen over the baseline. This is expected, as even random embeddings will provide semantic \textit{boundary} information to the MDE model, if not any useful semantic world knowledge. The authors emphasise that the backbone used in this method is considerably smaller than that used in the SOTA methods (see section \ref{sec:mde-model}), and thus we show results compared to a similarly-sized baseline.

When introducing different language embeddings, a further improvement in performance is seen in most cases. The best-performing model used the BERT-tiny embeddings, which have a dimensionality of only 128 channels compared to the 768 of BERT-base. This is surprising, as BERT-base outperforms BERT-tiny in many common natural language tasks \cite{turc_well-read_2019}, and it would be reasonable to assume that, if a depth bias exists in language, then it would be more clearly modelled by BERT-base than BERT-tiny. The second-best performing model, however, used the BERT-base embeddings. It may be the case that the size of the BERT-base embeddings presents too complex a mapping problem for the comparatively simple L2D models to fully learn, which would explain this observation.

Another notable observation is that the "inst"-pretrained L2D models outperformed the "loo" L2D models that used similar training and embedding size most of the time. The nature of the "leave-one-out" pretraining setup required that the resultant L2D models be frozen; this precludes the possibility of the model continuing to refine its predictions as training progresses. Further to this, the "inst"-trained models receive more information about the underlying NYUD2 dataset, seeing each instance as a datapoint instead of pre-aggregated information about all instances together. This may allow the model to learn more complex information about the nature of the depth-distribution bias encoded by the language models.

\section{Conclusion And Future Work}

In this work, we have shown that a depth-bias exists in natural language, and demonstrated two different ways of extracting this bias from BERT and BERT-tiny using simple auxiliary models. A novel method of incorporating these language-to-depth bias extraction models to an off-the-shelf monocular depth estimation pipeline was demonstrated, and an improvement in performance compared to both the baseline (no language embedding) and control (random language "embeddings") on the NYUD2 dataset was observed. Our method is simple to train and apply, and is extensible: any language embedding and any depth estimation model may be used with minimal modification, yielding a simple and accessible improvement on MDE performance.

Future work will explore the nature of language model depth biases further. In particular, our experiments have shown that certain configurations of language-to-depth model perform better than others at extracting BERT's depth bias. Therefore, the enhancement of these language-to-depth models to better extract depth bias and other depth-related world knowledge will be explored.

{\small
\bibliographystyle{ieee_fullname}
\bibliography{references}

\begin{thebibliography}{10}\itemsep=-1pt

\bibitem{auty_monocular_2022}
Dylan Auty and Krystian Mikolajczyk.
\newblock Monocular {Depth} {Estimation} {Using} {Cues} {Inspired} by
  {Biological} {Vision} {Systems}.
\newblock In {\em International {Conference} on {Pattern} {Recognition}
  ({ICPR}) 2022}, Apr. 2022.
\newblock arXiv: 2204.10384.

\bibitem{bai_monocular_2019}
Yucai Bai, Lei Fan, Ziyu Pan, and Long Chen.
\newblock Monocular {Outdoor} {Semantic} {Mapping} with a {Multi}-task
  {Network}.
\newblock {\em arXiv pre-print}, Jan. 2019.
\newblock arXiv: 1901.05807.

\bibitem{bhat_adabins_2020}
Shariq~Farooq Bhat, Ibraheem Alhashim, and Peter Wonka.
\newblock {AdaBins}: {Depth} {Estimation} using {Adaptive} {Bins}.
\newblock {\em arXiv:2011.14141 [cs]}, Nov. 2020.
\newblock arXiv: 2011.14141.

\bibitem{deng_imagenet:_2009}
Jia Deng, Wei Dong, Richard Socher, Li-Jia Li, Kai Li, and Li Fei-Fei.
\newblock {ImageNet}: {A} {Large}-{Scale} {Hierarchical} {Image} {Database}.
\newblock In {\em {CVPR}}, 2009.

\bibitem{devlin_bert_2019}
Jacob Devlin, Ming-Wei Chang, Kenton Lee, and Kristina Toutanova.
\newblock {BERT}: {Pre}-training of {Deep} {Bidirectional} {Transformers} for
  {Language} {Understanding}.
\newblock {\em arXiv:1810.04805 [cs]}, May 2019.
\newblock arXiv: 1810.04805.

\bibitem{dosovitskiy_image_2021}
Alexey Dosovitskiy, Lucas Beyer, Alexander Kolesnikov, Dirk Weissenborn,
  Xiaohua Zhai, Thomas Unterthiner, Mostafa Dehghani, Matthias Minderer, Georg
  Heigold, Sylvain Gelly, Jakob Uszkoreit, and Neil Houlsby.
\newblock An {Image} is {Worth} 16x16 {Words}: {Transformers} for {Image}
  {Recognition} at {Scale}.
\newblock {\em arXiv preprint}, page~21, 2021.

\bibitem{eigen_predicting_2015}
David Eigen and Rob Fergus.
\newblock Predicting {Depth}, {Surface} {Normals} and {Semantic} {Labels} with
  a {Common} {Multi}-{Scale} {Convolutional} {Architecture}.
\newblock In {\em {ICCV}}, 2015.

\bibitem{eigen_depth_2014}
David Eigen, Christian Puhrsch, and Rob Fergus.
\newblock Depth {Map} {Prediction} from a {Single} {Image} using a
  {Multi}-{Scale} {Deep} {Network}.
\newblock In {\em {NIPS}}, 2014.

\bibitem{ettinger_what_2020}
Allyson Ettinger.
\newblock What {BERT} is not: {Lessons} from a new suite of psycholinguistic
  diagnostics for language models, July 2020.
\newblock arXiv:1907.13528 [cs].

\bibitem{fu_deep_2018}
Huan Fu, Mingming Gong, Chaohui Wang, Kayhan Batmanghelich, and Dacheng Tao.
\newblock Deep {Ordinal} {Regression} {Network} for {Monocular} {Depth}
  {Estimation}.
\newblock In {\em {CVPR}}, 2018.

\bibitem{harkness_chameleons_1977}
Lindesay Harkness.
\newblock Chameleons use accommodation cues to judge distance.
\newblock {\em Nature}, 267:346--349, 1977.

\bibitem{he_deep_2016}
Kaiming He, Xiangyu Zhang, Shaoqing Ren, and Jian Sun.
\newblock Deep {Residual} {Learning} for {Image} {Recognition}.
\newblock In {\em {CVPR}}, 2016.
\newblock arXiv: 1512.03385v1.

\bibitem{henlein_what_2022}
Alexander Henlein and Alexander Mehler.
\newblock What do {Toothbrushes} do in the {Kitchen}? {How} {Transformers}
  {Think} our {World} is {Structured}.
\newblock In {\em Proceedings of the 2022 {Annual} {Conference} of the {North}
  {American} {Chapter} of the {Association} for {Computational} {Linguistics}
  ({NAACL} 2022)}, Apr. 2022.
\newblock Number: arXiv:2204.05673 arXiv:2204.05673 [cs].

\bibitem{hershenson_pictorial_1998}
Maurice Hershenson.
\newblock Pictorial {Cues}, {Oculomotor} {Adjustments}, {Automatic}
  {Organizing} {Processes}, and {Observer} {Tendencies}.
\newblock In {\em Visual {Space} {Perception}: {A} {Primer}}, pages 87--105.
  MIT Press, 1998.

\bibitem{hewitt_structural_2019}
John Hewitt and Christopher~D Manning.
\newblock A {Structural} {Probe} for {Finding} {Syntax} in {Word}
  {Representations}.
\newblock In {\em Proceedings of the 2019 {Annual} {Conference} of the {North}
  {American} {Chapter} of the {Association} for {Computational} {Linguistics}
  ({NAACL} 2019)}, volume~1, page~10, Minneapolis, Minnesota, 2019.

\bibitem{jiao_look_2018}
Jianbo Jiao, Ying Cao, Yibing Song, and Rynson Lau.
\newblock Look {Deeper} into {Depth}: {Monocular} {Depth} {Estimation} with
  {Semantic} {Booster} and {Attention}-{Driven} {Loss}.
\newblock In {\em {ECCV}}, 2018.

\bibitem{lee_big_2019}
Jin~Han Lee, Myung-Kyu Han, Dong~Wook Ko, and Il~Hong Suh.
\newblock From {Big} to {Small}: {Multi}-{Scale} {Local} {Planar} {Guidance}
  for {Monocular} {Depth} {Estimation}.
\newblock {\em arXiv:1907.10326 [cs]}, Aug. 2019.
\newblock arXiv: 1907.10326.

\bibitem{li_depthformer_2022}
Zhenyu Li, Zehui Chen, Xianming Liu, and Junjun Jiang.
\newblock {DepthFormer}: {Exploiting} {Long}-{Range} {Correlation} and {Local}
  {Information} for {Accurate} {Monocular} {Depth} {Estimation}.
\newblock {\em arXiv:2203.14211 [cs]}, Mar. 2022.
\newblock arXiv: 2203.14211.

\bibitem{li_binsformer_nodate}
Zhenyu Li, Xuyang Wang, Xianming Liu, and Junjun Jiang.
\newblock {BinsFormer}: {Revisiting} {Adaptive} {Bins} for {Monocular} {Depth}
  {Estimation}.
\newblock {\em arXiv preprint}, page~21, 2022.

\bibitem{liu_swin_2021}
Ze Liu, Yutong Lin, Yue Cao, Han Hu, Yixuan Wei, Zheng Zhang, Stephen Lin, and
  Baining Guo.
\newblock Swin {Transformer}: {Hierarchical} {Vision} {Transformer} using
  {Shifted} {Windows}.
\newblock {\em arXiv:2103.14030 [cs]}, Mar. 2021.
\newblock arXiv: 2103.14030.

\bibitem{may_measuring_2019}
Chandler May, Alex Wang, Shikha Bordia, Samuel~R. Bowman, and Rachel Rudinger.
\newblock On {Measuring} {Social} {Biases} in {Sentence} {Encoders}.
\newblock In {\em Proceedings of the 2019 {Conference} of the {North}}, pages
  622--628, Minneapolis, Minnesota, 2019. Association for Computational
  Linguistics.

\bibitem{mikolov_efficient_2013}
Tomas Mikolov, Kai Chen, Greg Corrado, and Jeffrey Dean.
\newblock Efficient {Estimation} of {Word} {Representations} in {Vector}
  {Space}, Sept. 2013.
\newblock arXiv:1301.3781 [cs].

\bibitem{nagata_depth_2012}
Takashi Nagata, Mitsumasa Koyanagi, Hisao Tsukamoto, Shinjiro Saeki, Kunio
  Isono, Yoshinori Shichida, Fumio Tokunaga, Michiyo Kinoshita, Kentaro
  Arikawa, and Akihisa Terakita.
\newblock Depth {Perception} from {Image} {Defocus} in a {Jumping} {Spider}.
\newblock {\em Science}, 335:469, 2012.

\bibitem{pennington_glove_2014}
Jeffrey Pennington, Richard Socher, and Christopher Manning.
\newblock Glove: {Global} {Vectors} for {Word} {Representation}.
\newblock In {\em Proceedings of the 2014 {Conference} on {Empirical} {Methods}
  in {Natural} {Language} {Processing} ({EMNLP})}, pages 1532--1543, Doha,
  Qatar, 2014. Association for Computational Linguistics.

\bibitem{petroni_language_2019}
Fabio Petroni, Tim Rocktäschel, Sebastian Riedel, Patrick Lewis, Anton
  Bakhtin, Yuxiang Wu, and Alexander Miller.
\newblock Language {Models} as {Knowledge} {Bases}?
\newblock In {\em Proceedings of the 2019 {Conference} on {Empirical} {Methods}
  in {Natural} {Language} {Processing} and the 9th {International} {Joint}
  {Conference} on {Natural} {Language} {Processing} ({EMNLP}-{IJCNLP})}, pages
  2463--2473, Hong Kong, China, 2019. Association for Computational
  Linguistics.

\bibitem{radford_learning_2021}
Alec Radford, Jong~Wook Kim, Chris Hallacy, Aditya Ramesh, Gabriel Goh,
  Sandhini Agarwal, Girish Sastry, Amanda Askell, Pamela Mishkin, Jack Clark,
  Gretchen Krueger, and Ilya Sutskever.
\newblock Learning {Transferable} {Visual} {Models} {From} {Natural} {Language}
  {Supervision}.
\newblock {\em arXiv:2103.00020 [cs]}, Feb. 2021.
\newblock arXiv: 2103.00020.

\bibitem{ramamonjisoa_sharpnet:_2019}
Michaël Ramamonjisoa and Vincent Lepetit.
\newblock {SharpNet}: {Fast} and {Accurate} {Recovery} of {Occluding}
  {Contours} in {Monocular} {Depth} {Estimation}.
\newblock {\em arXiv preprint}, 2019.
\newblock arXiv: 1905.08598v1.

\bibitem{ramesh_hierarchical_2022}
Aditya Ramesh, Prafulla Dhariwal, Alex Nichol, Casey Chu, and Mark Chen.
\newblock Hierarchical {Text}-{Conditional} {Image} {Generation} with {CLIP}
  {Latents}.
\newblock {\em arXiv:2204.06125 [cs]}, Apr. 2022.
\newblock arXiv: 2204.06125.

\bibitem{simonyan_very_2015}
Karen Simonyan and Andrew Zisserman.
\newblock Very {Deep} {Convolutional} {Networks} for {Large}-{Scale} {Image}
  {Recognition}.
\newblock In {\em {ICLR}}, 2015.
\newblock arXiv: 1409.1556v6.

\bibitem{sousa_judging_2011}
Rita Sousa, Eli Brenner, and Jeroen B.~J. Smeets.
\newblock Judging an unfamiliar object's distance from its retinal image size.
\newblock {\em Journal of Vision}, 11(9):10--10, Aug. 2011.
\newblock Publisher: The Association for Research in Vision and Ophthalmology.

\bibitem{tan_efficientnet_2019}
Mingxing Tan and Quoc~V. Le.
\newblock {EfficientNet}: {Rethinking} {Model} {Scaling} for {Convolutional}
  {Neural} {Networks}.
\newblock In {\em Proceedings of the 36th {International} {Conference} on
  {Machine} {Learning}}, 2019.
\newblock arXiv: 1905.11946.

\bibitem{tjong_kim_sang_introduction_2003}
Erik~F. Tjong Kim~Sang and Fien De~Meulder.
\newblock Introduction to the {CoNLL}-2003 {Shared} {Task}:
  {Language}-{Independent} {Named} {Entity} {Recognition}.
\newblock In {\em Proceedings of the {Seventh} {Conference} on {Natural}
  {Language} {Learning} at {HLT}-{NAACL} 2003}, pages 142--147, 2003.

\bibitem{turc_well-read_2019}
Iulia Turc, Ming-Wei Chang, Kenton Lee, and Kristina Toutanova.
\newblock Well-{Read} {Students} {Learn} {Better}: {On} the {Importance} of
  {Pre}-training {Compact} {Models}.
\newblock {\em arXiv:1908.08962 [cs]}, Sept. 2019.
\newblock arXiv: 1908.08962.

\bibitem{vaswani_attention_2017}
Ashish Vaswani, Noam Shazeer, Niki Parmar, Jakob Uszkoreit, Llion Jones,
  Aidan~N. Gomez, Lukasz Kaiser, and Illia Polosukhin.
\newblock Attention {Is} {All} {You} {Need}.
\newblock {\em arXiv:1706.03762 [cs]}, Dec. 2017.
\newblock arXiv: 1706.03762.

\bibitem{wagner_barn_1991}
Hermann Wagner and Frank Schaeffel.
\newblock Barn owls ({Tyto} alba) use accommodation as a distance cue.
\newblock {\em Journal of Comparative Physiology A}, 169:515--521, 1991.

\bibitem{wang_language_2020}
Chenguang Wang, Xiao Liu, and Dawn Song.
\newblock Language {Models} are {Open} {Knowledge} {Graphs}.
\newblock {\em arXiv:2010.11967 [cs]}, Oct. 2020.
\newblock arXiv: 2010.11967.

\bibitem{zhang_can_2022}
Renrui Zhang, Ziyao Zeng, and Ziyu Guo.
\newblock Can {Language} {Understand} {Depth}?
\newblock In {\em {ACM} {Multimedia} 2022}. arXiv, July 2022.
\newblock arXiv:2207.01077 [cs].

\bibitem{zhou_scene_2017}
Bolei Zhou, Hang Zhao, Xavier Puig, Sanja Fidler, Adela Barriuso, and Antonio
  Torralba.
\newblock Scene {Parsing} {Through} {ADE20K} {Dataset}.
\newblock In {\em {CVPR}}, page~9, 2017.

\bibitem{zhou_semantic_2018}
Bolei Zhou, Hang Zhao, Xavier Puig, Tete Xiao, Sanja Fidler, Adela Barriuso,
  and Antonio Torralba.
\newblock Semantic {Understanding} of {Scenes} through the {ADE20K} {Dataset}.
\newblock {\em arXiv:1608.05442 [cs]}, Oct. 2018.
\newblock arXiv: 1608.05442.

\end{thebibliography}
}

\end{document}